\documentclass[letterpaper, 10 pt, conference]{ieeeconf}  

\IEEEoverridecommandlockouts                              

\usepackage{todonotes}
\usepackage{graphicx}
\usepackage{times} 
\usepackage{amsmath} 
\usepackage{amssymb}  

\usepackage{algorithm}
\usepackage{algpseudocode}
\usepackage{booktabs}
\newcommand{\prm}[1]{\textsc{#1}}
\title{\LARGE \bf
GAVEL: Graph World Models for Verified and Efficient \\ Long-Horizon LLM Task Planning
}

\author{Ruiyang Wang, Hao-Lun Hsu, Swarajh Mehta, Jiwoo Kim, Zhihao Dou, Miroslav Pajic}

\begin{document}

\maketitle
\thispagestyle{empty}
\pagestyle{empty}


\begin{abstract}
Large language models (LLMs) provide a flexible interface for long-horizon robot planning, but generated plans often fail to respect embodiment constraints, recover from planning errors, or reason effectively under partial observability. We present \textbf{GAVEL}, a framework for verifying and repairing long-horizon LLM planning built around an explicit \emph{graph world model}. The graph represents relevant object-relations, action pre-conditions and effects, and probabilistic beliefs over unobserved object locations. This model can predict the consequences of LLM-generated actions before execution, detect violations, and repair those whose corrections follow directly from the world model. This method also reserves LLM replanning solely for errors requiring semantic reasoning. For multi-task instructions, GAVEL reasons over distributions of possible object locations to reorder remaining subtasks and minimize expected search cost. We evaluate GAVEL on BEHAVIOR-1K across 100 single long-horizon tasks and 500 multi-task instructions. With Qwen3-8B, GAVEL improves single-task success from 41.2\% to 91.8\% and multi-task success from 19.9\% to 92.6\%. Distributional belief reasoning also reduces travel distance by approximately 5.4\% compared with a static variant. These improvements show that an explicit graph world model harness can substantially improve the reliability and efficiency of long-horizon embodied planning across compact local and frontier hosted LLM capabilities.
\end{abstract}


\section{Introduction}
\label{sec:introduction}
Large language models (LLMs) offer a promising approach for translating natural-language instructions into a long-horizon sequence of robot actions~\cite{ahn2022can,huang2022inner,liang2023code,song2023llm}. However, the resulting plans 
are not always executable:~they may omit prerequisite actions, violate embodiment constraints, or end without reaching the intended state. 
It was shown that LLMs are unreliable as standalone planners~\cite{valmeekam2023planning}, 
with failure rates that grow with planning horizon, particularly with compact models~intended for 
edge deployment. This motivates pairing language semantics with an explicit model that can verify a plan before~execution~\cite{kambhampati2024llms}.

Scene-graph methods (e.g., SayPlan~\cite{rana2023sayplan}) adopt a \emph{propose-and-check} loop: a structured environment model symbolically executes the candidate plan, and infeasible actions are returned to the LLM for another attempt. Yet, recovering from an error does not always require invoking an LLM.~For example, when a grasp fails for insufficient proximity,  the appropriate recovery action follows directly from the violated precondition. Since the graph that detects the failure already determines its correction, calling an LLM raises cost~and risks introducing additional errors. 

Such localized repairs are exactly the operations used by EPoG~\cite{yang2026integrated}-style planners, which synthesize entire plans from graph edits between the current and goal scene graphs. However, while graph-edit planning is fast and reliable for spatial rearrangement, it cannot express complex semantic goals (e.g.,~``wash~the~plate'') 
that require a sequence of actions, rather than  a simple change in the object relations (e.g., EPoG only supports pick and place).

This motivates treating the graph not as a passive verifier, but as a \emph{world model}: 
one that simulates the outcome of each action and applies the repairs 
implied by its modeled state transitions, 
reserving the LLM call for failures that genuinely require semantic reasoning.
Unlike a verifier, such a model could also capture what the robot does not yet know. A robot may know an environment's layout while remaining uncertain where task-relevant objects are; this matters most when an instruction contains several subtasks, since searching for one object updates the beliefs about the others and can change which task is cheapest to do next. Semantic priors can guide the search for unseen objects~\cite{ramakrishnan2022poni,yu2023l3mvn,ginting2024seek,wang2025comres} and belief-space planning adjusts as observations resolve uncertainty~\cite{garrett2020online,curtis2024partially}, but collapsing each belief to its most likely outcome and fixing task order at initialization leaves this information unused.

\begin{figure}[t]
    \centering
    \includegraphics[width=0.92\linewidth]{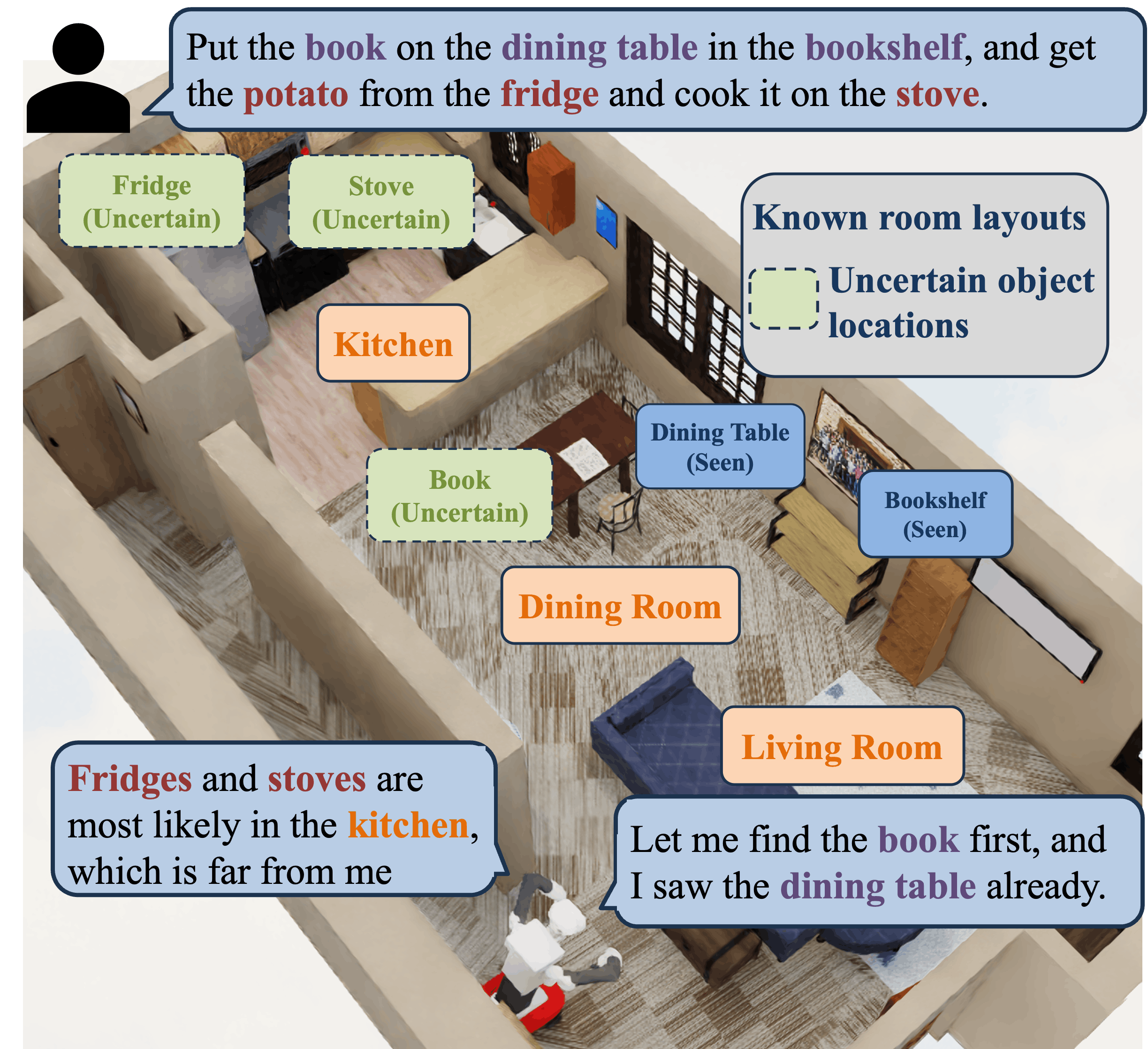}
    \caption{BEHAVIOR-1K~\cite{li2023behavior} example for long-horizon planning under partial observability. GAVEL verifies and repairs inconsistent LLM plans and uses object-location beliefs to reorder tasks online (i.e., during execution).
    }
    \label{fig:teaser}
\end{figure}

Consequently, this work introduces \textbf{GAVEL}: \textbf{G}r\textbf{a}ph World Models for \textbf{V}erified and \textbf{E}fficient Long-Horizon \textbf{L}LM Planning. GAVEL uses one explicit graph world model as the substrate for semantic LLM planning, symbolic consequence reasoning, and probabilistic environment beliefs. Given an LLM-generated plan, GAVEL rolls its actions forward on the graph, verifies preconditions, goal completion, and safety constraints, and directly repairs model-derivable failures. Under partial observability, the graph maintains distributions over the locations of unseen objects and converts them into expected search and navigation costs. After each completed task, new observations update these beliefs and GAVEL reoptimizes the remaining task order. 

Specifically, the main contributions of this work are:

\begin{itemize}
\item A graph world model that verifies LLM plans against action preconditions, goal completion, and safety. It repairs the failures its own action semantics determine, and passes only semantic failures back to the LLM.

\item An extension to partially observed multi-task execution that retains full distributions over unobserved object locations, converts them into expected search cost, and re-optimizes the remaining task order online.

\item An evaluation on BEHAVIOR-1K~\cite{li2023behavior} across single long-horizon tasks, and partially observed multi-task instructions. On tests with both local and hosted LLMs, we demonstrate substantial improvements in task success and execution efficiency, and show that explicit world-model reasoning remains complementary to increasing LLM capability.
\end{itemize}

\section{Related Work}
\label{sec:related}

\subsection{LLMs for Embodied Task Planning}

LLMs are increasingly used as high-level planners for embodied agents. SayCan~\cite{ahn2022can} scores candidate actions by learned affordances, Code as Policies~\cite{liang2023code} and ProgPrompt~\cite{singh2022progprompt} have the LLM write executable policies directly, and LLM-Planner~\cite{song2023llm} and Inner Monologue~\cite{huang2022inner} replan from observed or reported feedback. A second group pairs the LLM with explicit solvers instead. LLM+P~\cite{liu2023llm+} formulates the problem for a classical planner, while PRoC3S~\cite{curtis2024trust} and Text2Motion~\cite{lin2023text2motion} check constraint satisfaction before execution. The prior works consistently show that an explicit model's feedback improves the  reliability of task planning.

\subsection{Scene-Graph Grounding and Verification}
Graph structures have recently been explored as explicit world representations for agent reasoning, from a general Graph World Model for structured and multimodal prediction~\cite{feng2025graph} to AriGraph~\cite{anokhin2024arigraph}, which learns a knowledge-graph world model with semantic and episodic memory for LLM agents. In robotics, scene graphs offer a natural planning substrate by grounding objects, spatial relations, and states directly~\cite{agia2022taskography,jiao2022sequential}. SayPlan~\cite{rana2023sayplan} and VeriGraph~\cite{ekpo2024verigraph} use them only for pre-execution verification, and LookPlanGraph~\cite{onishchenko2025lookplangraph} refines them online via VLMs. GAVEL instead treats the graph as a full world model, representing action transitions and uncertainty for subsequent planning.

\subsection{Graph Planning under Partial Observability}

Scene graphs can serve as persistent representations as an environment is incrementally observed~\cite{amiri2022reasoning,rajvanshi2024saynav,honerkamp2024language}. 
When task-relevant objects remain unseen, semantic knowledge can guide where to search. PONI~\cite{ramakrishnan2022poni} learns semantic potential functions over the unexplored space, SEEK~\cite{ginting2024seek} combines a dynamic scene graph with a Relational Semantic Network (RSN) to predict likely object locations, and L3MVN~\cite{yu2023l3mvn} and COMRES-VLM~\cite{wang2025comres} use LLM/VLM commonsense to guide exploration. GAVEL follows SEEK's RSN formulation for initializing room-location beliefs, while retaining the resulting distribution for downstream planning.

EPoG~\cite{yang2026integrated} is the closest baseline to GAVEL under
partial observability. It maintains a belief graph of observed and predicted objects and constructs the global plan directly from graph edits toward the goal, using an LLM mainly for situated local replanning as observations arrive. GAVEL reverses this division of responsibility: the LLM generates the semantic procedure, while the graph acts as an explicit \emph{world model} that rolls the plan forward, detects violations, and directly repairs failures whose corrections are implied by the modeled action semantics. Only failures requiring additional semantic reasoning are returned to the LLM. GAVEL further retains full distributions over unobserved object locations and uses them to estimate execution costs and reoptimize task order online. This connects GAVEL to prior work on belief-space replanning~\cite{garrett2020online,curtis2024partially} and multi-task
subgoal selection~\cite{wang2023describe}.

\section{Problem Definition}
\subsection{Problem Statement}
\label{sec:statement}

Consider a robot in an indoor environment with room instances $\mathcal{R}$ of semantic type $\tau:\mathcal{R}\rightarrow\mathcal{T}$, where $\mathcal{T}$ is the set of semantic types (e.g., \{kitchen, dining room, etc.\}), and task-relevant objects $\mathcal{O}$ with categories $c(o)$. We represent the environment by a typed scene graph defined as:
\begin{equation}
G=(V,E,\Phi), \qquad V=\mathcal{R}\cup\mathcal{O}\cup\{\rho\},
\end{equation}
where $\rho$ is the robot, $E$ represents relations including $\{$\emph{near}, \emph{under}, \emph{room\_connect}, \emph{room\_inside}, \emph{object\_inside}, \emph{on\_top}, \emph{next\_to}, \emph{holding}$\}$, and $\Phi$ contains the unary flags $\{\mathrm{open}, \mathrm{toggled}, \mathrm{cooked}, \mathrm{washed}, \mathrm{dried}\}$. We distinguish the hidden physical state $G_t^\star$ from the robot's believed graph $G_t$, for execution step $t\in\{0,\ldots,H-1\}$.

We assume that the room layout $(\mathcal{R},\tau)$ and its connectivity are known, and object relations specified by the instruction or already observed are represented deterministically in $G_t$. The room locations of remaining unobserved objects are uncertain. Let $x_t\in\mathcal{R}$ denote the robot's room at step $t$.

The robot acts through grounded primitives $a_t=(\alpha,\xi)\in\mathcal{A}$, where $\alpha$ is the action type and $\xi$ its argument. Each primitive has graph-valued preconditions and effects $\langle \mathrm{pre}_a,\mathrm{eff}_a\rangle$ defining the transition model
\begin{equation}
f(G,a)=
\begin{cases}
\mathrm{eff}_a(G), & \text{if } \mathrm{pre}_a(G) \text{ holds},\\
\bot, & \text{otherwise},
\end{cases}
\label{eq:process}
\end{equation}
where $\bot$ denotes an invalid transition. A plan $\pi=\langle a_0,\ldots,a_{H-1}\rangle$ is \emph{applicable} from $G_0$ if every action satisfies its preconditions along the induced trace. GAVEL uses~\eqref{eq:process} as its predictive world model, while physical execution evolves the hidden state $G_t^\star$.
The environment is partially observed, so we maintain a belief over the room locations of uncertain objects,
\begin{equation}
B_t=\{b_o^t\}_{o\in\mathcal{O}}, \qquad
b_o^t(r)=P\left(\mathrm{room}(o)=r \mid \mathcal{H}_t\right),
\label{eq:belief}
\end{equation}
where $\mathcal{H}_t$ is the history of actions and observations up to step $t$. The observation at each step, $z_t$, is updated through the sensor model $\Xi$ as:
$z_t=\Xi(x_t,G_t^\star).$

For each uncertain object, the observation either localizes it, if present, or prunes the observed room from its candidate set if absent. The belief is updated and actions are selected by an execution policy $\mu$ from the belief state,
\begin{equation*}
B_{t+1}=\Upsilon(B_t,a_t,z_{t+1}), \quad a_t = \mu(G_t,B_t),
\end{equation*}
while $G_t^\star$ is revealed only through observations.

An instruction $\mathcal{I}$ is decomposed into $N$ tasks $\{\mathcal{I}_i\}_{i=1}^{N}$ with partial-graph goals $\{g_i\}_{i=1}^{N}$, each specifying relations and unary states that must hold at termination. The overall goal is $g=\bigcup_{i=1}^{N} g_i$, satisfied when $\mathsf{unmet}(g,G)=\emptyset$. We additionally impose a trace-level safety predicate $\mathsf{safe}(a_{0:H-1})$, capturing restoration requirements such as closing opened containers and switching off safety-critical appliances.

We consider multi-task instructions whose tasks are independent by construction. Their goal-object sets are disjoint, $\mathrm{obj}(g_i)\cap\mathrm{obj}(g_j)=\emptyset$ for $i\neq j$, so every ordering is admissible and ordering affects execution cost rather than goal feasibility. Generated subplans may nevertheless interact through auxiliary objects or state changes, motivating explicit validation of their composition.

\textit{Problem:} \textbf{Verified Multi-Task Planning under Partial Observability.}
\label{prob:main}
Given an instruction $\mathcal{I}$, an initial believed graph $G_0$ with belief $B_0$, and an unknown initial physical state $G_0^\star$ consistent with $B_0$, find a policy $\mu \in \mathcal{M}$ minimizing the expected physical execution distance, where
$\mathcal{M}$ denotes the set of admissible policies of the form $\mu(G_t,B_t) \in \mathcal{A}$:
\begin{equation}
\mu^\star =
\arg\min_{\mu\in\mathcal{M}}
\mathbb{E}_{G^\star_0\sim B_0,\mu}
\left[C_{\mathrm{exec}}(\mu)\right].
\label{eq:objective}
\end{equation}
where
\begin{equation}
C_{\mathrm{exec}}(\mu) = \sum_{t=0}^{H-1}\ell(a_t;G_t^\star),
\end{equation}
subject to
\begin{align}
&\mathsf{unmet}(g,G_H^\star) = \emptyset,
& \mathsf{safe}(a_{0:H-1}) \text{ holds}, \nonumber\\
& \mathrm{pre}_{a_t}(G_t^\star) \text{ holds},
& G_{t+1}^\star = f(G_t^\star,a_t), \nonumber\\
& a_t = \mu(G_t,B_t),
& z_{t+1} = \Xi(x_{t+1},G_{t+1}^\star), \nonumber\\
& B_{t+1} = \Upsilon(B_t,a_t,z_{t+1}). \nonumber
\end{align}
Here $\ell(a_t;G_t^\star)$ is the physical navigation distance incurred by $a_t$, and the constraints hold for all $t\in\{0,\ldots,H-1\}$.

Problem~\ref{prob:main} couples two difficulties. First, an LLM-generated plan is not guaranteed to satisfy the transition constraints in~\eqref{eq:process}, so executability must be checked explicitly. Second, execution cost depends on uncertain object locations and changes as observations update $B_t$. GAVEL addresses both through a common graph world model: explicit action semantics support verification and repair, while the evolving belief supports uncertainty-aware task ordering.

\subsection{Belief-Aware Task Ordering}
\label{sec:objective}

For a multi-task instruction, GAVEL selects remaining task execution order to minimize expected travel under current belief $B_t$. For an unlocalized object $o$, let $(r_1,\ldots,r_K)$ denote its reachable candidate rooms, ordered by decreasing $b_o^t(r)$ where shortest-path distance breaks ties. If $o$ is found in the $k$-th searched room, the incurred search cost is
\begin{equation*}
\kappa_k =
D(x_t,r_1)
+\sum_{m=1}^{k-1}
\left[s(r_m)+D(r_m,r_{m+1})\right]
+\eta s(r_k),
\end{equation*}
where $D(\cdot,\cdot)$ is the shortest-path distance,
$s(r)=A_{\mathrm{trav}}(r)/w_{\mathrm{cov}}$ approximates the full search cost of room $r$, $A_{\mathrm{trav}}(r)$ is the traversable area of room $r$, $w_{\mathrm{cov}}$ is the effective sensor coverage width, and $\eta\in[0,1]$ is the expected fraction searched before detection. The expected cost of reaching an unlocalized object is
\begin{equation*}
C_{\mathrm{srch}}(x_t,o;B_t)
=
\frac{1}{Z_t}
\sum_{k=1}^{K} b_o^t(r_k)\kappa_k,
\qquad
Z_t=\sum_{k=1}^{K} b_o^t(r_k).
\label{eq:csearch}
\end{equation*}
If $o$ is localized, we use its geometric navigation cost,
\begin{equation}
C_{\mathrm{loc}}(q_t,o)
=
d_{\mathrm{nav}}(q_t,p_o),
\end{equation}
where $q_t$ and $p_o$ denote the robot and object positions. These costs are used during belief-conditioned rollouts to estimate the cost of executing each task and transitioning between.

Specifically, let $h[j]$ denote the expected cost of executing task $j$ first from the current state $(x_t,B_t)$, and let $A[i,j]$ denote
the expected cost of executing task $j$ after task $i$. For an ordering $\varsigma=(\varsigma_1,\ldots,\varsigma_N)$, we approximate its
expected execution cost by
\begin{equation}
J(\varsigma)
=
h[\varsigma_1]
+
\sum_{p=2}^{N}
A[\varsigma_{p-1},\varsigma_p].
\label{eq:J}
\end{equation}
Constructing $(h,A)$ requires $\mathcal{O}(N^2)$ belief-conditioned rollouts. GAVEL executes only the first task of the minimum-cost ordering, updates $B_t$ using the resulting observations, and recomputes the ordering for the remaining tasks. Since $N\leq5$ in our experiments, all candidate permutations are enumerated and Eq.~\eqref{eq:J} is minimized exactly.

\section{Method}
\label{sec:method}

\subsection{System Overview}
\label{sec:overview}
\begin{figure*}[ht!]
    \centering
    \includegraphics[width=1.0\linewidth]{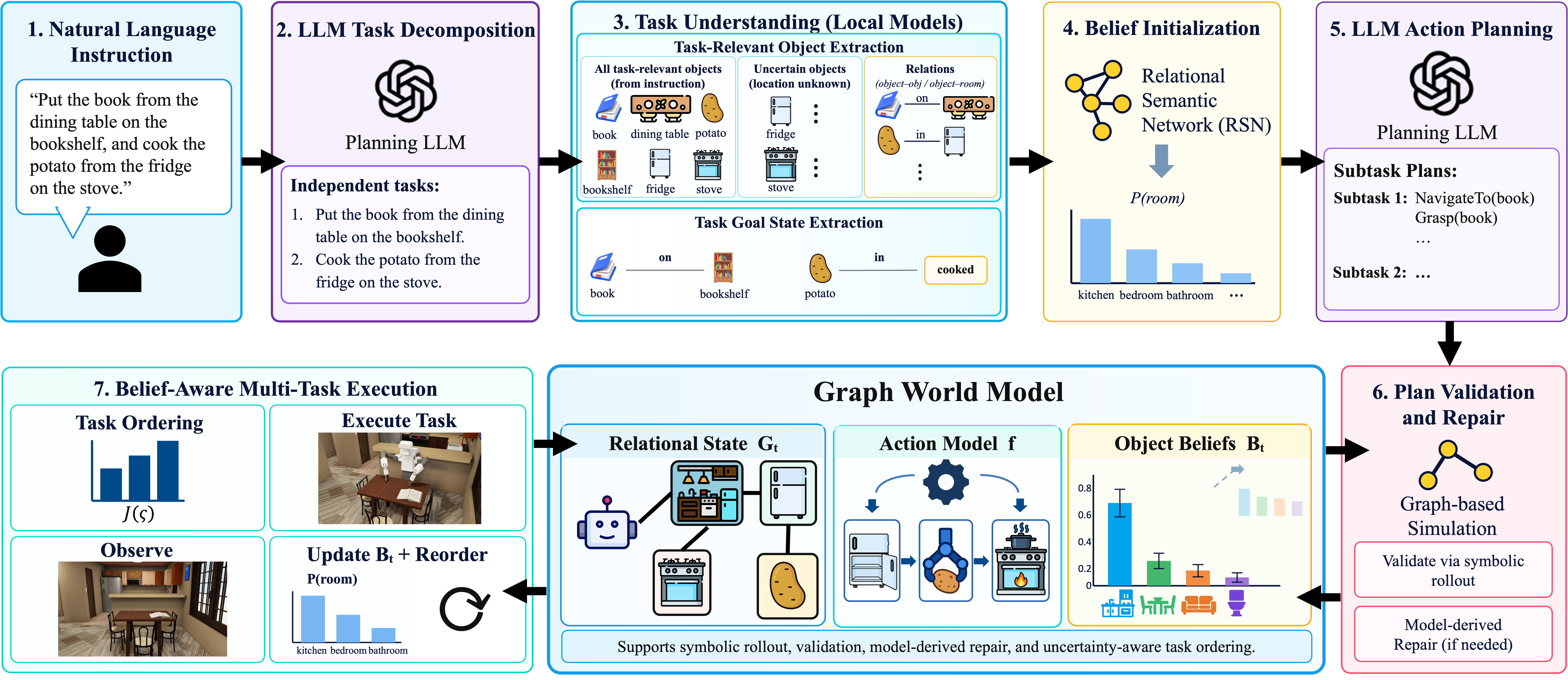}
    \caption{\small Overview of the GAVEL pipeline. The instruction is decomposed into tasks, objects, and extracted goals. Uncertain objects receive RSN predicted location beliefs. An LLM query generates task plans, which the graph world model verifies and repairs before execution. GAVEL then updates beliefs from observations and reorders the remaining tasks online.
    }
    \label{fig:pipeline}
\end{figure*}

GAVEL combines semantic LLM planning with an explicit graph world model for verified, belief-aware long-horizon execution. As shown in Fig.~\ref{fig:pipeline} and Alg.~\ref{alg:gavel}, an instruction $\mathcal{I}$ is first decomposed into tasks $\{\mathcal{I}_i\}_{i=1}^{N}$. Task understanding extracts the entities, relations, and goals needed to initialize the graph, while the RSN assigns room-location beliefs to uncertain objects.

A single LLM query then produces an action plan $\pi_i$ per task. The graph world model rolls each plan forward, repairs model-derivable failures, and returns unresolved violations to the LLM at most $T$ (e.g., 5) times. Once verified, GAVEL orders the plans by the cost in Eq.~\eqref{eq:J} under the current belief. After each completed task, new observations update the belief and the remaining order is reoptimized.

\begin{algorithm}[t]
\caption{GAVEL}
\label{alg:gavel}
\begin{algorithmic}[1]
\Require instruction $\mathcal{I}$; believed graph $G_0$; LLM budget $T$

\State $\{\mathcal{I}_i\}_{i=1}^{N}\gets\textsc{Decompose}(\mathcal{I})$
\For{$i=1$ \textbf{to} $N$}
    \State $(\mathcal{O}_i,\mathcal{E}_i)\gets
    \textsc{ExtractObjects}(\mathcal{I}_i)$
    \State $\mathcal{O}_i^{\mathrm{qry}}\gets
    \textsc{Uncertain}(\mathcal{O}_i,\mathcal{E}_i)$
    \State $g_i\gets\textsc{ParseGoal}(\mathcal{I}_i)$ \quad $G_0\gets\textsc{Insert}(G_0,\mathcal{E}_i)$
\EndFor
\State $B_0\gets
\textsc{RSN}(\bigcup_i\mathcal{O}_i^{\mathrm{qry}},G_0)$

\State $p\gets\textsc{Prompt}(\mathcal{I},G_0,B_0,\{g_i\})$
\For{$iter=1$ \textbf{to} $T$}
    \State $\{\pi_i\}_{i=1}^{N}\gets
    \textsc{Parse}(\textsc{LLM}(p))$
    \For{$i=1$ \textbf{to} $N$}
        \State $\pi_i\gets\textsc{Repair}(G_0,\pi_i,g_i)$
        \State $\omega_i\gets\textsc{Validate}(G_0,\pi_i,g_i)$
    \EndFor
    \If{all $\omega_i$ are valid}
        \State \textbf{break}
    \EndIf
    \State $p\gets\textsc{Complaint}
    (\mathcal{I},G_0,\{\pi_i,\omega_i\})$
\EndFor

\State $\mathcal{U}\gets\{1,\ldots,N\}$,\quad
       $(G,B)\gets(G_0,B_0)$
\While{$\mathcal{U}\neq\emptyset$}
    \State $(h,A)\gets\textsc{Rollout}(\mathcal{U},B,x)$
    \State $\Sigma\gets
    \textsc{Sort}_{J}\!\left(\mathrm{Perm}(\mathcal{U})\right)$
    \For{$\varsigma\in\varSigma$}
        \State $\omega\gets\textsc{Validate}
        \!\left(
        G,\,
        \bigoplus_{j=1}^{|\mathcal{U}|}\pi_{\varsigma_j},\,
        \bigcup_{i\in\mathcal{U}}g_i
        \right)$
        \If{$\omega$ is valid}
            \State \textbf{break}
        \EndIf
    \EndFor
    \State $(G,z)\gets\textsc{Execute}(\pi_{\varsigma_1})$
    \State $B\gets\Upsilon(B,\pi_{\varsigma_1},z)$
    \quad $\mathcal{U}\gets\mathcal{U}\setminus\{\varsigma_1\}$
\EndWhile
\end{algorithmic}
\end{algorithm}

\subsection{Task Understanding}
\label{sec:task_understanding}

For each decomposed task $\mathcal{I}_i$, GAVEL extracts (i) task-relevant objects and relations to initialize the belief graph and (ii) goal states used to evaluate task completion (Fig.~\ref{fig:pipeline}). Both mappings use lightweight LoRA adapters of Qwen3-1.7B. These are shared by all planner models and baselines.

\paragraph{Task-relevant object extraction}
The grounding adapter maps $\mathcal{I}_i$ to an object set $\mathcal{O}_i$ and explicitly stated relations $\mathcal{E}_i$, which are inserted directly into the believed graph rather than inferred from semantic priors. For example, ``get the potato from the fridge'' yields $\texttt{inside}(\texttt{potato},\texttt{fridge})$, localizing the potato through the fridge. Only objects whose locations remain uncertain are passed to the RSN:
\begin{equation}
\mathcal{O}_i^{\mathrm{qry}}
=
\{o\in\mathcal{O}_i:
\mathrm{room}(o)\text{ is uncertain in }\mathcal{E}_i\},
\end{equation}

\paragraph{Task-goal extraction}
A second adapter maps $\mathcal{I}_i$ to a partial-graph $g_i$ describing the goal terminal state. It contains spatial predicates like $\texttt{on\_top}(o_1,o_2)$ and $\texttt{inside}(o_1,o_2)$, and semantic
states like $\texttt{washed}(o)$, and $\texttt{dried}(o)$. 

Both adapters operate on natural-language instruction and are trained on $8{,}000$ synthetic instruction--target pairs with a disjoint $500$-instruction validation set. 

\subsection{Belief Initialization}
\label{sec:rsn}

For each uncertain object $o\in\bigcup_i\mathcal{O}_i^{\mathrm{qry}}$,
GAVEL initializes a room-location belief using a Relational Semantic Network (RSN), following SEEK~\cite{ginting2024seek}. A frozen text encoder $\phi(o)\in\mathbb{R}^{384}$ followed by an MLP $f_\theta$ predicts a score for each room type $\tau\in\mathcal{T}$:
\begin{equation}
o
\xrightarrow{\phi}
\phi(o)
\xrightarrow{f_\theta}
\{\tilde f_o(\tau)\}_{\tau\in\mathcal{T}}.
\label{eq:rsn}
\end{equation}

Because the RSN predicts room types while GAVEL reasons over room instances, each type score is distributed uniformly among reachable instances of that type,
\begin{equation}
b_o^0(r)
\propto
\frac{\tilde f_o(\tau(r))}
{\left|\{r'\in\mathcal{R}^{\mathrm{re}}:
\tau(r')=\tau(r)\}\right|}.
\label{eq:rsnbelief}
\end{equation}
The distribution is then normalized over reachable rooms.

We use \texttt{BAAI/bge-small-en-v1.5} as the frozen encoder and a three-layer MLP with hidden dimensions $256$--$128$--$64$ and dropout $0.2$. The RSN is trained from scene object placements using weighted binary cross-entropy and calibrated using Platt scaling. 

\subsection{LLM Action Planning}
\label{sec:llm_planning}

After task understanding and belief initialization, the planner receives the instruction $\mathcal{I}$, believed graph $G_0$, predicted object-location beliefs $B_0$, task goals $\{g_i\}_{i=1}^{N}$, and the available action definitions. As shown in Alg.~\ref{alg:gavel}, a single
LLM query produces one grounded action sequence for each decomposed task,
\begin{equation}
\{\pi_i\}_{i=1}^{N}
=
\mathrm{LLM}(\mathcal{I},G_0,B_0,\{g_i\}_{i=1}^{N}).
\end{equation}

Each resulting $\pi_i$ is then passed to the graph world model for validation and repair. If a violation cannot
be resolved from the modeled action semantics, its structured failure description is returned to the LLM for another planning attempt, up to a budget number of calls $T$.

\subsection{Graph World Model}
\label{sec:worldmodel}

The graph world model provides the symbolic transition used to
predict the consequences of LLM-generated actions. GAVEL operates over
nine grounded primitives,
\begin{equation*}
\begin{aligned}
&\mathcal{A}=\{
\prm{NavigateTo}, \prm{Grasp}, \prm{Release}, \prm{PlaceOnTop},\\
&\prm{PlaceInside}, \prm{Open}, \prm{Close}, \prm{ToggleOn}, \prm{ToggleOff}
\}.
\end{aligned}
\label{eq:actions}
\end{equation*}
Each grounded action $a=(\alpha,\xi)$ has modeled preconditions and effects $\langle\mathrm{pre}_a,\mathrm{eff}_a\rangle$, defining the transition in Eq.~\eqref{eq:process}.

The preconditions capture four forms of executability:
\emph{proximity}, requiring the robot to be near the interaction target; \emph{affordance}, requiring the target to support the requested action; \emph{gripper state}, enforcing holding constraints; and \emph{accessibility}, requiring enclosing containers to be open. \prm{NavigateTo} establishes proximity but cannot target an object hidden inside a closed container.

Action effects update graph relations and object states. \prm{Grasp} establishes the holding relation, placement transfers the held object to its destination, and appliance operation can induce semantic states such as $\mathrm{cooked}$, $\mathrm{washed}$, or $\mathrm{dried}$. GAVEL also tracks restoration constraints: containers opened during a plan must be closed, and safety-critical appliances switched on must later be switched off.

\subsection{Plan Validation and Repair}
\label{sec:validation_repair}

\begin{algorithm}[!t]
\caption{\textsc{Repair}: Model-derived plan repair}
\label{alg:repair}
\begin{algorithmic}[1]
\Require believed graph $G$, plan $\pi$, goal $g$
\State $\pi^\star \gets \pi$ \; $\omega^\star \gets \textsc{Validate}(G,\pi^\star,g)$ \; $\mathcal{S} \gets \{\pi\}$
\Loop
    \State $\omega \gets \textsc{Validate}(G,\pi,g)$
    \If{$\mathrm{rk}(\omega,\pi) > \mathrm{rk}(\omega^\star,\pi^\star)$}
        \State $(\pi^\star,\omega^\star) \gets (\pi,\omega)$
    \EndIf
    \If{$\omega.\mathrm{app}$ and $\omega.\mathrm{safe}$ and $\omega.\mathrm{unmet}=\emptyset$}
        \State \Return $\pi$
    \EndIf
    \If{no model-derived edit exists for $\omega$}
        \State \textbf{break}
    \EndIf
    \State $\pi' \gets \textsc{Edit}(\pi,\omega,g)$
    \If{$\pi'\in\mathcal{S}$}
        \State \textbf{break}
    \EndIf
    \State $\mathcal{S}\gets\mathcal{S}\cup\{\pi'\}$ \quad $\pi\gets\pi'$
\EndLoop
\State \Return $\pi^\star$
\end{algorithmic}
\end{algorithm}

Given a candidate plan $\pi_i=\langle a_0,\ldots,a_{H-1}\rangle$, GAVEL predicts its consequences by rolling the actions forward on the graph world model. Each action is applied only if its preconditions hold. Otherwise rollout stops at the first violation. The resulting graph is then checked against goal $g_i$ and trace-level safety constraints. A plan is valid iff
\begin{equation}
\mathsf{unmet}(g_i,G_H)=\emptyset
\quad\text{and}\quad
\mathsf{safe}(\pi_i).
\end{equation}
The validator returns a structured verdict $\omega$ identifying applicability failures, unmet goal conditions, and safety violations.

Many such failures imply their own corrections. As summarized in Alg.~\ref{alg:repair}, \textsc{Repair} recursively applies edits whose effects follow directly from the action model: a failed grasp due to missing proximity induces \prm{NavigateTo}, and an object inside a closed container induces navigation to the container followed by \prm{Open}. Safety violations similarly induce restoration actions. When no model-derived repair exists, the unresolved verdict is returned to the LLM as structured feedback. The visited set $\mathcal{S}$ prevents cyclic repair by ending when an edit leads to a previously considered plan.

Because successive edits may not monotonically improve a plan, GAVEL retains the best previous candidate:
\begin{equation*}
\mathrm{rk}(\omega,\pi)
=
\left(
\mathbf{1}[\omega.\mathrm{app}],
\mathbf{1}[\omega.\mathrm{unmet}=\emptyset],
\mathbf{1}[\omega.\mathrm{safe}],
-\omega.d
\right).
\label{eq:rk}
\end{equation*}
where $\omega.d$ defines the execution cost of the plan. Thus, the budget $T$ limits LLM generations, while graph-derived repair may perform multiple edits between two LLM queries.

\subsection{Belief-Aware Multi-Task Execution}
\label{sec:ordering}

Once the task plans are repaired and verified, GAVEL determines which remaining task to execute next under current belief $B_t$. At each task boundary, \textsc{Rollout} constructs the first-task costs $h$ and pairwise transition costs $A$ using the cost model in Sec.~\ref{sec:objective}. Eq.~\eqref{eq:J} evaluates every permutation of the remaining task set $\mathcal{U}$. GAVEL sorts these orderings by increasing expected cost and validates each one's concatenated plans against the union of the corresponding goals, selecting the first that passes. If the cheapest ordering causes cross-task interference, the next is considered instead.

Only the first task of the selected ordering is executed. The resulting observations update the beliefs of all uncertain objects: observed objects become localized, while searched rooms in which an object is not seen are removed from its belief support. GAVEL then rebuilds $(h,A)$ and reoptimizes the remaining order, so information gathered during one task immediately affects the schedule of those that follow.

\section{Experimental Evaluation}
\label{sec:experiments}

\subsection{Simulation Setup}
\label{sec:exec}
We evaluate GAVEL at the symbolic action level of BEHAVIOR-1K, as our focus is high-level task planning rather than low-level manipulation. For large-scale evaluation, we implement the BEHAVIOR-1K action transitions in a lightweight 2-D executor retaining the original floor plans, object properties, task predicates, and navigation geometry. Navigation is geometric: traversability maps are eroded by the robot base radius and paths are planned with A$^\star$. Unseen objects are searched through frontier-based exploration with a wedge-shaped camera. A completed room search localizes observed objects and rules that room
out for unobserved ones. For the ordering cost model, we use
$w_{\mathrm{cov}}=1.2\,\mathrm{m}$. The action space is unchanged: all nine primitives map one-to-one onto the BEHAVIOR-1K/OmniGibson interface, and every benchmark task is directly executable in OmniGibson. To validate the abstraction, we ran all $100$ single-task instructions in OmniGibson; success outcomes agreed with the executor on every instance. The executor reduces execution time from roughly $17$ minutes per plan to about one second, enabling $5{,}500$ evaluations at scale. The supplementary video shows GAVEL executing example tasks in OmniGibson.

We measure \emph{Success} with the executor after the plan is simulated, \emph{Distance} as how far the robot travels, and \emph{Planning time} as decomposition, LLM calls, repair, and ordering, measured on an idle GPU. Results are averaged over five random seeds and reported as mean $\pm$ std.

\subsection{Benchmarks}
\label{sec:bench}
Prior work in this setting typically evaluates on relatively small task sets: five object-transport tasks across 46 scenes in EPoG~\cite{yang2026integrated}, and 90 instructions in SayPlan~\cite{rana2023sayplan}. We follow the same evaluation paradigm at substantially larger scale, first verifying every reference plan in our executor so measured failures reflect planning rather than invalid specifications.

The \textbf{single-task benchmark} holds $100$ long-horizon tasks across $10$ scenes: fetching objects, opening containers, operating appliances, and delivering things to a specified destination. Reference plans average $12.2$ primitives ($8$--$17$) and collectively include $240$ goal predicates.
The \textbf{multi-task benchmark} holds $500$ instructions over the 
scenes, each with $N\in\{2,3,4,5\}$ independent tasks,~averaging $18.8$ primitives across $4.66$ rooms. Of these, $199$ need a semantic state that only operating an appliance can produce.

\subsection{Compared Methods}
\label{sec:baselines}
Every method sees the same initial state, LLM budget, and believed graph; the SayPlan and EPoG baselines reproduce their relevant mechanism inside this shared pipeline.

\texttt{llm-only} generates one plan and runs it. \texttt{sayplan} adds the validation-feedback loop of~\cite{rana2023sayplan} with up to $T$ LLM re-calls, isolating \emph{feedback without repair}. \texttt{epog} derives actions from the difference between the believed and goal graphs, ordered by travel cost~\cite{yang2026integrated}, isolating planning from graph-state differences alone. \texttt{oracle} uses ground-truth locations and reference subplans as a cost reference.

Among our variants, \texttt{gavel-basic} takes one LLM proposal and then relies only on graph repair, isolating \emph{repair without feedback}. \texttt{gavel-map} commits each uncertain object to its most likely room and fixes the order up front; \texttt{gavel-static} keeps the full distribution but still fixes the order. Finally, \texttt{gavel} keeps the distribution and reorders after every finished task.

\subsection{Belief and Cost-Model Quality}
\label{sec:beliefquality}
The RSN trains on $11{,}218$ placements across $51$ scenes, $197$ object categories, and $37$ room types, with scenes held out. It reaches an AUC of $0.904$ and a Brier score of $0.057$. Still, the most likely room is a poor summary of that belief: across $832$ held-out queries the true room ranks first only $47\%$ of the time, so collapsing $b_o$ to $\hat r_o$ discards information.

The pairwise cost model tracks what the robot actually drives, at $R^2=0.93$ for four-task and $R^2=0.89$ for five-task instructions, which is what justifies the $\mathcal{O}(N^2)$ construction instead of rolling out every permutation. 
\subsection{Experiment 1: Single Long-Horizon Task Reliability}
\label{sec:exp1}

\begin{table}[t]
\centering
\caption{Experiment~1: 100 single long-horizon tasks.}
\label{tab:exp1}
\footnotesize
\setlength{\tabcolsep}{2pt}
\begin{tabular}{@{}llccc@{}}
\toprule
Model & Method & Success (\%) & Average Attempt & Time (s) \\
\midrule
Qwen3-4B & \texttt{llm-only} & $21.8\pm1.5$ & $1.00 \pm 0.0$ & $9.2\pm1.0$ \\
Qwen3-4B & \texttt{sayplan} & $55.4\pm1.9$ & $3.4\pm1.7$ & $27.6\pm12.0$ \\
Qwen3-4B & \texttt{gavel-basic} & $67.7\pm1.6$ & $1.00\pm0.0$ & $9.2\pm1.0$ \\
Qwen3-4B & \texttt{gavel} & $\mathbf{88.8} \pm \mathbf{1.6}$ & $1.7\pm1.3$ & $15.0\pm11.1$ \\
\midrule
Qwen3-8B & \texttt{llm-only} & $41.2\pm1.2$ & $1.00\pm0.0$ & $13.3\pm6.4$ \\
Qwen3-8B & \texttt{sayplan} & $76.4\pm1.5$ & $2.7\pm1.7$ & $24.6\pm15.8$ \\
Qwen3-8B & \texttt{gavel-basic} & $76.2\pm1.2$ & $1.00\pm0.0$ & $13.5\pm6.4$ \\
Qwen3-8B & \texttt{gavel} & $\mathbf{91.8} \pm \mathbf{1.3}$ & $1.5\pm1.1$ & $15.5\pm9.4$ \\
\bottomrule
\end{tabular}
\end{table}

We first ask whether an LLM can reliably generate a complete long-horizon plan before considering partial observability or multi-task scheduling. As shown in Table~\ref{tab:exp1}, \textbf{LLM-only} planning is highly unreliable for the compact models: Qwen3-4B and Qwen3-8B succeed on only $21.8\%$ and $41.2\%$ of tasks, confirming that simply generating a plan and executing it is insufficient at these horizons.

A common alternative is to detect an invalid plan and return the failure to the LLM for replanning, as in SayPlan and related feedback-based approaches~\cite{rana2023sayplan,huang2022inner}. This improves success to $55.4\%$ and $76.4\%$, but requires $3.4$ and $2.7$ LLM calls per task. More importantly, feedback does not guarantee that a compact LLM will correctly repair the reported error.

\begin{figure}
    \centering
    \includegraphics[width=1.0\linewidth]{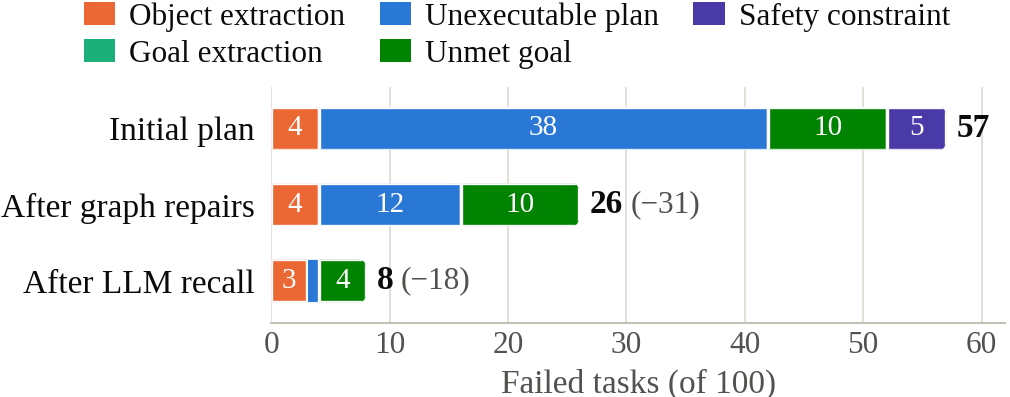}
    \caption{\small
    Failure breakdown for Qwen3-8B GAVEL in Experiment~1. The figure shows remaining failures after each pipeline step.
}
    \label{fig:failure}
\end{figure}

\textbf{Graph-derived repair} addresses this limitation directly. Consider \texttt{gavel-basic}, a variant that only performs graph repairs and never re-queries the LLM. It already reaches $67.7\%$ with Qwen3-4B and $76.2\%$ with Qwen3-8B. On Qwen3-4B this exceeds feedback-only replanning by $12.3$ points using $1$ rather than $3.4$ queries, and on Qwen3-8B it matches it at the same cost. When GAVEL additionally returns only the remaining semantic failures to the LLM, success rises to $88.8\%$ and $91.8\%$ with just $1.7$ and $1.5$ calls on average. The graph thus handles failures implied by action semantics, while LLM replanning is reserved for those that genuinely require semantic reasoning. For Qwen3-4B this also cuts planning time from $27.6$\,s under feedback-only replanning to $15.0$\,s. Fig.~\ref{fig:failure} shows the failure breakdown for a representative Qwen3-8B run, illustrating how errors are removed across the pipeline and which remain.

\subsection{Experiment 2: Long-Horizon Multi-Task Execution}
\label{sec:exp2}

\begin{table}[t]
\centering
\caption{Experiment~2: 500 multi-task instructions using Qwen3-8B.}
\label{tab:multi}
\footnotesize
\setlength{\tabcolsep}{4pt}
\begin{tabular}{@{}lccc@{}}
\toprule
Method & Success (\%) & Distance (m) & Time (s) \\
\midrule
\texttt{llm-only} & $19.9\pm0.7$ & $86.69\pm0.37$ & $31.8\pm10.5$ \\
\texttt{sayplan} & $75.6\pm2.1$ & $83.01\pm0.42$ & $51.4\pm27.6$ \\
\texttt{epog} & $60.2\pm0.0$ & $84.51\pm0.00$ & $13.9\pm3.9$ \\
\texttt{gavel-map} & $\mathbf{92.6} \pm \mathbf{1.0}$ & $82.45\pm0.23$ & $39.3\pm22.6$ \\
\texttt{gavel-static} & $\mathbf{92.6} \pm \mathbf{1.0}$& $79.69\pm0.45$ & $39.3\pm22.6$ \\
\texttt{gavel} & $\mathbf{92.6} \pm \mathbf{1.0}$ & $\mathbf{78.01} \pm \mathbf{0.71}$ & $39.3\pm22.6$ \\
\midrule
\texttt{oracle} & $100.0 \pm 0.0$ & $56.17 \pm 0.00$ & --- \\
\bottomrule
\end{tabular}
\end{table}

Having established reliable single-task planning, we next consider long-horizon instructions containing multiple tasks under partial observability. Here failures compound: although each component task is shorter than in Experiment~1 ($4.86$ versus $12.2$ reference primitives), the complete instruction's success is dependent on all component plans. As a result, \texttt{llm-only} completes just $19.9\%$ instructions.

\textbf{Verification and repair} remain important at this scale. The SayPlan-style baseline, which returns graph-validation failures to the LLM but performs no graph-derived repair, reaches $75.6\%$; GAVEL completes $92.6\%$. GAVEL also reduces traveled distance from $83.01$\,m to $78.01$\,m on the matched sets both methods succeeded, and planning time by roughly $12$\,s per instruction. Repeatedly asking the LLM is therefore not equivalent to directly correcting failures the action model has already determined.

The EPoG-style baseline exposes the complementary limitation of \textit{graph-only procedure generation}. It plans~by~transforming the believed graph toward the goal graph and succeeds on $60.2\%$ of instructions. Its failures correspond exactly to the instructions requiring an appliance-induced semantic state (e.g., \texttt{cooked}, \texttt{washed}, or \texttt{dried}); ``wash~the plate'' requires inferring that the dishwasher must be run, a procedure the graph-state difference alone cannot specify. This motivates keeping the LLM as the high-level semantic planner, while the graph world model determines if the LLM's procedure is executable and repairs 
inconsistencies.

GAVEL's own $38$ residual failures on Experiment 2 lie mostly upstream of planning: $21$ due to object extraction, with only $5$ unexecutable plans, and $9$ unmet goals. 

Finally, we isolate GAVEL's uncertainty reasoning. \textbf{Distributional beliefs} reduce mean distance from $82.45$\,m with \texttt{gavel-map} to $79.69$\,m with \texttt{gavel-static}, and \textbf{online reordering} after each completed task further reduces it to $78.01$\,m. Together these save $4.45$\,m, or $\approx 5.4\%$, while the ordering optimization itself takes only $17$\,ms per instruction.

\subsection{Experiment 3: Does Model Scaling Replace GAVEL?}
\label{sec:exp3}

Finally, we ask if the above gains could instead be obtained 
by replacing compact local models with stronger 
LLMs. We evaluate all planners on the same 100-instruction subset of the multi-task benchmark, selecting every fifth instruction.

\begin{table}[t]
\centering
\caption{Experiment~3: local and hosted planners on the same
100-instruction subset.}
\label{tab:exp3}
\setlength{\tabcolsep}{2.5pt}
\begin{tabular}{@{}lcccc@{}}
\toprule
& \multicolumn{2}{c}{Success (\%)}
& \multicolumn{2}{c}{Time (s)} \\
\cmidrule(lr){2-3}
\cmidrule(lr){4-5}
Planner & LLM & GAVEL & LLM & GAVEL \\
\midrule
Qwen3-4B
& $2.4\pm1.5$ & $\mathbf{75.2}\pm\mathbf{2.1}$
& $23.0\pm7.4$ & $35.0\pm31.2$ \\
Qwen3-8B
& $23.6\pm1.5$ & $\mathbf{89.2}\pm\mathbf{1.3}$
& $31.8\pm10.5$ & $39.3\pm22.6$ \\
GPT-5.6 Sol
& $24.6\pm3.9$ & $\mathbf{99.2}\pm\mathbf{0.8}$
& $16.5\pm3.9$ & $16.5\pm4.0$ \\
Claude Sonnet 5
& $38.6\pm2.2$ & $\mathbf{99.4}\pm\mathbf{0.5}$
& $18.0\pm4.4$ & $18.2\pm4.4$ \\
\bottomrule
\end{tabular}
\end{table}

\textbf{Scaling alone} improves semantic planning but does not eliminate long-horizon failures. Without GAVEL, Qwen3-4B and Qwen3-8B solve only $2.4\%$ and $23.6\%$ of the subset, while GPT-5.6 Sol and Claude Sonnet~5 reach $24.6\%$ and $38.6\%$. In contrast, Qwen3-4B with GAVEL reaches $75.2\%$, substantially outperforming the hosted models used alone. The gain from explicit world-model reasoning therefore seems to persist despite increasing LLM capability.

\textbf{Scaling and GAVEL are complementary}. With GAVEL, Qwen3-8B reaches $89.2\%$ and both hosted models reach near-perfect success on this subset. Stronger LLMs reduce the remaining semantic planning errors, while GAVEL removes the action-level applicability and consistency failures that persist even for frontier models.

This is also inexpensive for the hosted models: planning time remains $16.5$\,s for GPT-5.6 Sol and changes only from $18.0$ to $18.2$\,s for Claude Sonnet~5. The additional cost for compact models therefore comes from further LLM planning attempts rather than from graph verification and repair.

\section{Conclusion}
We presented GAVEL, a framework for long-horizon LLM planning built around an explicit graph world model representing object relations, action preconditions and effects, and beliefs over unobserved object locations. GAVEL rolls LLM-generated plans forward before execution, repairs failures whose corrections follow from the modeled action semantics, and reserves LLM queries for failures requiring semantic reasoning. Under partial observability, it uses distributional beliefs to estimate execution costs and reoptimizes the remaining task order as observations arrive.

Our results highlight the benefit of this division of responsibility. On BEHAVIOR-1K, graph-derived repair alone raises Qwen3-4B single-task success to $67.7\%$, exceeding feedback-only replanning. Allowing LLM re-queries for unresolved semantic failures further raises this to $88.8\%$. World-model reasoning is also complementary to scaling: Qwen3-4B with GAVEL outperforms stronger hosted planners when used alone, while both hosted models reach near-perfect success when combined with GAVEL. Retaining full object-location distributions and reoptimizing online further reduces traveled distance by approximately $5.4\%$.

Finally, our evaluation uses symbolic manipulation, so the graph captures logical action consequences but not geometric feasibility. Adding motion-level reachability and collision constraints would extend the same verify-and-repair mechanism to physical execution. The semantic prior is conditioned primarily on room type, and improving it based on repeated observations could capture environment-specific regularities. Finally, most residual failures now arise in grounding the instruction rather than planning over the graph, pointing to instruction understanding as the next bottleneck.

\bibliographystyle{IEEEtran}
\bibliography{bib}

\end{document}